\documentclass[paper=a4wide, fontsize=12pt]{scrartcl}	 
\usepackage[a4paper, left=20mm, right=20mm, top=20mm, bottom=2.5cm, footskip=1.2cm]{geometry}  

\usepackage[svgnames]{xcolor} 
\usepackage{background} 
\usepackage{fancyhdr} 
\usepackage{graphicx}
\usepackage{ragged2e, setspace}
\usepackage[utf8]{inputenc}
\usepackage{caption,subcaption}

\usepackage{biblatex}
\usepackage{amsmath}
\usepackage{braket}

\backgroundsetup{contents={\includegraphics[width=\textwidth]{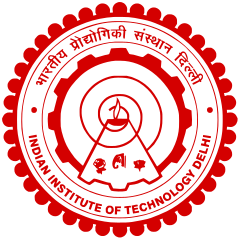}},scale=0.2,placement=bottom,opacity=0.4,position={7.2cm,50cm}} 

\title{\vspace{-1.8cm}  \color{DarkRed} On Fulfilling the Exigent Need for Automating and Modernizing Logistics Infrastructure in India}
\subtitle{Enabling AI-based Integration, Digitalization, and Smart Automation of Industrial Parks and Robotic Warehouses
}
\author{Shaurya Shriyam\footnote{Corresponding author: shriyam@iitd.ac.in} , Prashant Palkar, Amber Srivastava \\ Department of Mechanical \& Industrial Engineering, I.I.T. Delhi
\vspace{-2cm} }
\date{} 

\begin{document}

\begin{titlepage}
    \centering

    \includegraphics[width=17cm]{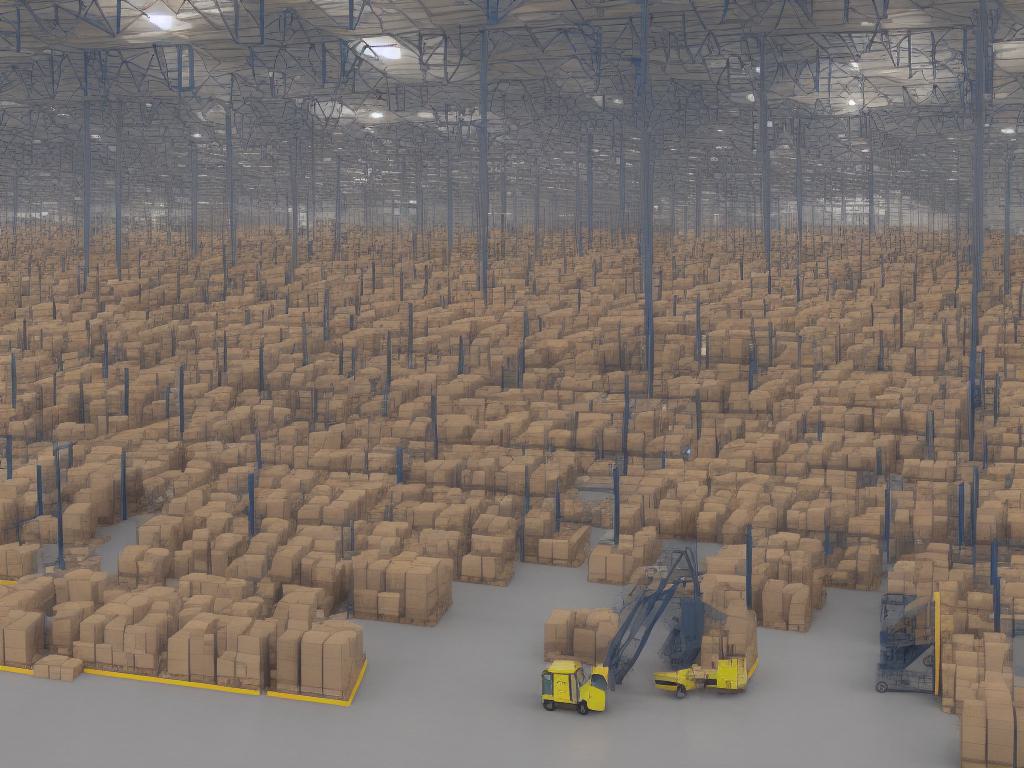}
    \noindent\textcolor{Goldenrod}{\rule{17cm}{1mm}}
    \vspace{0.5cm}
    \begin{spacing}{1.75}
    {\justify \bfseries \Huge ON FULFILLING THE EXIGENT NEED FOR AUTOMATING AND MODERNIZING LOGISTICS INFRASTRUCTURE IN INDIA}
    \end{spacing}
    \vspace{0.2cm}
    \begin{spacing}{1.5}
    {\Large
        Enabling AI-based Integration, Digitalization, and Smart Automation of Industrial Parks and Robotic Warehouses}
    \end{spacing}

\end{titlepage}

\maketitle


\vspace{0.5cm}

\section*{Executive Summary}

In this article, we emphasize the need for Low- or Middle-Income Countries (LMICs) to embrace Industry 4.0 and Logistics 4.0 to stay competitive. This requires government-level interventions and policy-making to incentivize quality product solutions and drive innovation in traditionally resistant economic sectors. We propose the establishment of Smart Industrial Parks (SIPs) with a focus on enhancing operational efficiencies and bringing together MSMEs and startups targeting niche clientele with innovative Industry 4.0 solutions. SIPs along with the phased deployment of well-planned robotic automation technologies shall enable bringing down India's untenable logistics costs. We discuss the concept of a Supply Hub in an Industrial Park (SHIP) as a solution to efficiently utilize land inside industrial parks for manufacturing and logistics purposes. The execution of SIPs faces challenges, particularly in competitive markets with fluctuating demand and time-sensitive customer demands. To tackle this, we are required to implement the efficient allocation of manufacturing resources and capabilities within SIPs. Thus, we emphasize the importance of efficient resource utilization, collaboration, and technology adoption in industrial parks to promote industrial development and economic growth. We advocate the use of a cloud-based cyber-physical system for real-time data access and analysis in SIPs. Such centralized cloud-based monitoring of factory floors, warehouses, and industrial units using IoT infrastructure shall improve decision-making, efficiency, and safety. Digital Twins (DTs), which are cyber-replicas of physical systems, could play a significant role in enabling simulation, optimization, and real-time monitoring of smart manufacturing and distributed manufacturing systems. However, there are several challenges involved in implementing DTs in distributed manufacturing systems, such as defining data schemas and collaboration protocols, ensuring interoperability, the need for effective authentication technology, distributed machine learning models, and scalability to manage multiple DTs. Our proposed vision to address these difficulties toward the establishment of Smart Industrial Parks is completely in line with the National Logistics Policy (NLP) and Prime Minister Gati Shakti National Master Plan for Multi-modal Connectivity and is also strongly related to the National Smart Cities Mission.

\section*{Introduction}

As society undergoes numerous technological advancements while tackling emerging social challenges that are characteristic of the LMICs in the Global South, it produces exigent circumstances requiring every operational aspect of human society to be transformed in order to adapt to the changing world around us. The backbone of any aspirational country's economy, the production and manufacturing sectors, have also witnessed a sea change. The vision of Industry $4.0$ encompasses the major ways in which this industrial sector is transforming by addressing individualization and decentralization of products, flexibility, and scalability of production systems while keeping up with sustainability goals. The main building blocks of Industry $4.0$ are Cyber-Physical Systems (CPS) and the Internet of Things (IoT). In the age of Industry $4.0$, automation, artificial intelligence, and IoT seamlessly integrate with manufacturing processes in order to improve their efficiency and ultimately produce AI-enabled smarter products. \\

However, simply focusing on innovations in the manufacturing process is not enough. The optimal utilization of the country's resources and realized profitability hinges on implementing efficient logistics infrastructure nationwide. Such a reciprocal vision for the logistics sector is dubbed as Logistics $4.0$ \cite{Winkelhaus2019}. Logistics $4.0$ refers to developing logistics solutions with the objective of meeting mass product customization demands by utilizing new-age AI/ML technologies based on IoT and CPS frameworks while not increasing incurred costs and meeting sustainability goals. Logistics $4.0$ is characterized by deploying real-time big data analytics for optimization, autonomous robots for better inventory management, and seamless error-free lightning-speed information exchange. Although Logistics $4.0/$Industry $4.0$ promises to boost profit margins along with increased growth opportunities, many industrial enterprises appear to be hesitant in transforming their operations because of the significant efforts required for the same and uncertainty about the technical know-how to achieve this goal. \\

Although Indian manufacturers have increased investment towards the vision of industry $4.0$ by $10$ times over a decade, we are still far behind meeting Industry $4.0$ standards. However, the adoption rate of Industry $4.0$ has recently increased because of the disruption caused by COVID-19. So, it is expected that by $2025$, about two-thirds of the Indian manufacturers will embrace Industry 4.0 standards. In $2021$, the total money spent by companies under the category of manufacturing technologies was $20\%$ and it is expected to increase by $40\%$ by $2025$. Automobile, electrical, electronics, chemical, and pharmaceutical companies are at the forefront of Industry $4.0$ adoption. It is common wisdom that in those economic sectors that are culturally resistant to change, the innovation mindset can be boosted by government-level interventions and policy-making that shift the focus away from investment costs and towards incentivizing quality product solutions that are internationally competitive  \cite{nasscom}. \\

In the context of Logistics $4.0$, the US-based Plus One Robotics (Texas, US) and Amazon Robotics have made excellent progress. Their much-touted warehouse automation solutions have significant implications for the Indian retail and logistics sector. Plus One Robotics provides robot perception solutions and logistics automation such as PickOne, which is a 3D and AI-powered vision software for induction, singulation, depalletizing, and packing. It has also developed Yonder, which is a remote supervisor software suite for robotic applications. The India-based Addverb Technologies delivers robotics solutions for material movement, order sequencing, picking and sorting, and Automated storage and retrieval systems (ASRS). It also develops warehouse management systems (WMSs) and robot fleet management systems (RMSs). Another India-based robotics company, Peer Robotics is bringing innovations to the field of lean manufacturing with its collaborative mobile robots. GreyOrange is another warehouse robotics company with an Indian-origin CEO that is delivering promising AI-based solutions. Another startup called GrayMatter Robotics founded in $2020$ by researchers in Prof. S.K. Gupta's lab has already clinched series A financing because of its efficient surface finishing automation services. \\

We know that maturity models are useful for businesses to measure how capable they are of continuous improvement. They are well-suited to provide a discrete roadmap for the transition process toward designing completely networked production. The initial maturity level appears to be mostly chaotic while lacking rules and process definitions. But in the final optimized level, all processes get defined, documented, standardized, analyzed, controlled, and optimized. In the context of Industry 4.0 maturity levels, the first level refers to an unconnected analog production. Then digital data processing is introduced at the next level whereas the third level of data collection is mostly automated. The fourth level is highly networked production and the final level is completely networked production \cite{Facchini2019}. \\

In the context of Logistics 4.0, we may define maturity levels for subsystems such as purchase logistics, production logistics, distribution logistics, and after-sales logistics. In each of these, maturity moves from no data exchange level to a completely automated data handling level. In general, the maturity levels for logistics could be assigned to be as follows: ignoring, defining, adopting, managing, and integrating. In the ``ignoring" level, people are mostly unaware of the Logistics 4.0 systems and in the ``defining" level people begin to realize the scope and need for such a system. The integration process begins at the ``adoption" level and if most of the systems become integrated, then we have reached the ``managed" level. In the final ``integrated" level, all the possible advanced solutions are integrated into the logistics operations \cite{Facchini2019}.

\section*{Smart Industrial Parks}

United Nations Industrial Development Organization (UNIDO) describes an Industrial park as a tract of land used by a large number of manufacturers for better cooperation and collaboration to provide cost-effective communal services. An industrial park is built by following structured plans for roads, transportation, logistics warehouses, advanced manufacturing facilities, and other resources that promote the overall industrial development and economic growth of the country \cite{QIU201316}. \\

To meet the rapidly increasing land demands for manufacturing, logistics related warehousing, and other set-ups present in industrial parks, an efficient utilization of the land available inside the industrial park is crucial. The concept of a Supply Hub in an Industrial Park (SHIP) helps in addressing this issue. SHIP is a concept of a public warehouse facility inside an industrial park for providing logistics and warehousing services. SHIP is a more sophisticated version of traditional distribution centers. It provides services like holding and management of inventory of raw materials as well as finished goods. It allows suppliers/manufacturers/customers/retailers to efficiently share the storage space originally owned by each enterprise. SHIP has advantages such as low per-unit transportation costs, efficient utilization of space inside the park, and expert handling and management of the inventory by third-party logistics providers. \\

However, it is proving to be very costly and challenging for MSMEs to shift towards Industry $4.0$. One major reason is that it is not economically feasible for MSMEs to spend their limited funds on robotics and AI because the scale at which these technologies start paying off is well beyond the reach of startups and niche businesses. Leaving behind the MSMEs and not making them a part of India's Logistics $4.0$ vision, which is at the heart of the National Logistics Policy, would be a grave miscalculation. Thus, a viable solution is to bring such small businesses under the umbrella of a Smart Industrial Park (SIP). An SIP is not much different than the vision of the Smart Cities Mission launched by the Government of India in $2015$. However, an SIP will be geared towards the objective of enhancing operational efficiencies by bringing together various industrial players in a collaborative fashion while operating at Technology Readiness Levels $7-9$. A standard industrial-cum-business park is built for the purpose of industrial development by bringing together offices, industries, and manufacturing units. Newer variants also foster the participation of startups and niche innovative businesses by incubating them and allowing them easier access to high-level industrial partners whose support they require intermittently to grow their business. Our proposal is along similar lines to bring together MSMEs and startups that are trying to cater to a niche clientele by targeting innovative solutions based on the latest technologies championed as part of Industry $4.0$, and give those MSMEs and startups access to more established industrial players who might be interested in helping these budding players in a mutually beneficial manner. The main theme of such SIPs would revolve around technologies such as AI/ML, IoT/CPS, robotics, and additive manufacturing. \\

However, the successful execution of SIP is fraught with challenges. In a competitive market with frequent product upgrades and fluctuating market demand, manufacturing businesses in an industrial park face highly time-sensitive and intensive customer demands. Manufacturing businesses are hardly able to fulfill customer demand under such challenging circumstances because of their constrained manufacturing resources and capabilities. This invariably results in the potential loss of customers to the companies. On the other hand, the companies might also experience a lack of customer orders because of a demand shortage, resulting in idle manufacturing resources and capabilities. \\

Cloud manufacturing (CMfg) offers to address this issue by making infrastructure for sharing manufacturing resources in an industrial park. In CMfg systems, organizations can share and use on-demand large-scale heterogeneous and geographically distributed manufacturing resources. The manufacturing capabilities are encapsulated via a cloud platform that provides Manufacturing Cloud Services (MCSs). However, to cut manufacturer-side costs, it is necessary that these services are allocated in an optimal manner. The entire success of a SIP hinges on the efficiency by which such an optimization is achieved. Auction-based mechanisms \cite{Kang2022} are well-suited to handle manufacturing cloud services allocation (MCSA) problems. \\

Thus, a group of manufacturers started using the MCSs inside an SIP to share resources and capabilities \cite{Kang2022}. To understand this, visualize this scenario. At any point during the operations in SIP, a set of manufacturers may end up having idle resources and capabilities (who may be referred to as providers) while other SIP participants may require manufacturing resources and capabilities (who may be referred to as customers). The providers and customers will keep changing with time. There can also be multiple ways to implement MCSs in such scenarios, so MCSs are typically diversified. Therefore, for any allocation of MCSs to be effective and optimal, all the types of MCSs should be taken into consideration. To resolve, this we could categorize the SIP's manufacturing capabilities and resources available into manufacturing clouds (MCs) and rank each service within an MC by measuring its Quality of service (QoS) level. Customers submit requests for obtaining MCSs to an appropriate QoS level MC while providers publish available MCSs corresponding to each MC when idle. The trades in different MCs shall function in a mutually independent manner. It is also conceivable to allow customers within the SIP or even an external entity to bid for more complicated operations involving multiple tasks subject to sequential constraints. For example, a requested operation may comprise multiple lathe machining, drilling, cutting, milling, and grinding tasks in a particular order. In such a case, customers would also be required to know from the providers the end-to-end quality of service support, and that too over heterogeneous networks if we assume that entities outside the SIP are also seeking collaboration. Amidst such complexity, it is challenging to devise an effective dynamic pricing strategy that ensures that the monetary payoffs of community cloud platform operators are maximized. \\

The SHIP supply chain consists of multiple suppliers, manufacturers, and customers within the same industrial park. Assuming that the manufacturer does not keep any inventory, the raw material is delivered to the SHIP from different suppliers and then consolidated to form the amount required by the manufacturers and then delivered together. Also, the different types of finished goods are consolidated at SHIP and delivered to the customers as per their requirements. The traditional supply chain also consists of multiple suppliers, manufacturers, and customers, but they do not have any intermediary between them. In the conventional supply chain without the SHIP, the raw material is directly supplied to the manufacturers by the different suppliers, and various manufacturers provide the finished goods directly to the customers. The manufacturers have to manage the raw material and finished goods inventory. To optimize the production-distribution supply chains, we must minimize the total cost of the manufacturers, including the cost of production, inventory holding, and transportation.

\section*{Enabling Warehouse Automation Using Digital Twins}

In smart manufacturing, collecting data with the help of IoT devices and analyzing the Big Data helps to discover the vulnerability of the existing system. It thereby leads to the optimization of the production system and the overall supply chain. Direct manipulation of the physical system may lead to disruptions. Thus, it is better to model a cyber replica of the system to simulate the changes. This replica is known as the Digital Twin (DT) and helps find the optimal modifications to be made in the system. The DT refers to a data-driven set of software and hardware tools that specifies an existing physical system along with all the functions and use cases \cite{fi14020064}. The adaptive version of such DTs helps intelligently predict the potential risks within the distributed manufacturing systems with the help of shared knowledge and real-time data. A better understanding of the potential risks facilitates consensus-based decision-making. \\

DTs are useful for predicting, controlling, or optimizing performance parameters while continuously learning from them. DTs share a live connection with the physical systems to monitor and update them in a proactive real-time manner. Apart from performing simulation, DTs also interact with the system to infer effective ways for it to adapt to environmental changes. DTs also make heavy use of advanced analytical tools such as Big Data analytics, machine learning, and deep learning to draw inferences from the massive amount of data obtained using IoT sensors and networking technologies. The critical point distinguishing DTs from traditional simulations is the extensive use of sensors, communication networks, data collection and analysis as well as cost-effective and intelligent decision-making modules to interpret the complex stochastic process and deliver real-time feedback to the system \cite{fi14020064}. \\

Any DT consists of three components: physical, digital, and the link between them. It is difficult to change the existing systems as they consist of complex operations. So, DT provides a simplified design of the current system and uses the continuous data from the system environment to perform simulations and guide the system on its next steps. The digital representation of an actual physical system along with a detailed description of its theoretical model is called the Digital Model (DM). Digital Shadow (DS) is an improvement over DM as it simulates and integrates a one-way data flow from the physical entities to the digital replica. Digital Twins (DTs) are the further extension of a digital shadow and digital model. A DT system connects the physical system with the digital model in a two-way manner. \\

The DT can also be divided into the following four categories \cite{fi14020064}: DT prototype (DTP), DT instance (DTI), DT aggregation (DTA), and DT environment (DTE). The DTP framework of the digital twins considers all the information from the existing system, which is essential to developing the cyber clone of the physical system. It improves the cost and time efficiency of the process of the actual system. Data flows one way from the physical to the cyber system. Thus, it is mainly used for monitoring the real system. Unlike DTP, DTI manages the data flow from the digital system to the physical one. The data flow includes forecasts or recommendations that help the physical system adapt to environmental changes. The two-way connection between the physical and digital systems can be established using DTP and DTI. DTI primarily provides a single data flow from the digital to the physical system. At the same time, DTA describes the combination of all DTIs. A DTE consists of multiple DT systems used to manage large systems. \\

Each physical system's IoT gateway gathers and combines data from the sensors. Communication protocols transmit data from the physical system to the AI model. Based on the conditions of physical systems, the AI model is utilized for data analysis and model construction. The AI model saves its learned model as a DT model, a replica of the existing system. If the physical system's configurations remain unchanged, the DT model may direct the physical system to respond depending on the information it has gathered. The AI model may update its model as per the changes to the physical system. It can rewrite the DT model to assist the physical system's operations further if the configuration of the physical system shifts or requires a different response to the same circumstance. \\

However, there are several challenges in implementing DT in distributed manufacturing systems \cite{machines9090193}. The DT's data schema, the collaboration protocols, and the policies for sharing internal and external data need to be defined to ensure interoperability. A digitally distributed manufacturing system requires effective authentication technology to secure real-time data exchange and analysis across many players. The massive data from different manufacturers leads to a requirement for an effective distributed machine learning model with accurate predictions. The distributed manufacturing system needs to accommodate a large number of DTs by being scalable and must manage multiple deployed DTs while simultaneously maintaining the requisite level of robustness. \\

The combined use of digital twins and blockchain technology has a significant potential to address these challenges. The main aim of combining these two techniques is to create a conceptual framework for data-driven ledger-based collaborative DT \cite{machines9090193}. The framework consists of a data-driven ledger-based predictive model, which uses DT-driven operational data to predict potential risks. The predictive model is crucial to building and assessing local DT deployment using DT-driven operational data. The collaboration aspect in decentralized DTs is enhanced via a distributed consensus algorithm. Several efficient distributed consensus methods are capable of obtaining a majority of nodes to agree on potential risks and alert the decision-makers in distributed manufacturing systems. \\

A cloud-based cyber-physical system allows access to real-time data from a remote location in the smart industrial park and helps in gaining insights by applying strategic data analysis algorithms. It helps foster better decision-making by increasing the efficiency, profit, and safety of the overall system. We propose to develop a centralized cloud-based monitoring of the smart industrial park that will comprise factory floors, warehouses, and industrial units monitored continuously using IoT-based infrastructure. The concept of SHIP we introduced above seamlessly integrates with this idea of centralized cloud-based monitoring of the SIP. \\

For a general manufacturing industry in a typical industrial park, industrial communication for the purpose of schedule updating or machine maintenance tasks happens via unofficial verbal meetings whose effectiveness is heavily dependent on worker expertise. Most companies will use these two types of classical decision-making and work prioritization techniques, First in First out (FIFO) and Earlier Due Date (EDD); and they would employ reactive maintenance strategies in the event of unforeseen machine-tool breakdowns. However, work output and industrial productivity could be significantly enhanced by adopting a cloud-based cyber-physical system, which will lead to more accurate, flexible, and adaptive task rescheduling and machine maintenance on the shop floor while handling uncertainties as well as unpredictable events. \\

A typical industry shop floor comprises job shops, which are defined as groups of work centers, which in turn are defined as groups of machines and include individual resources (machine tools \& operations). The shop floor is installed with Data Acquisition Systems (DAQs) and wireless sensor networks to capture data that are transmitted into MongoDB, the No-SQL database of the cloud server. This heterogeneous and high-volume data is analyzed by using the information fusion technique and relevant information such as availability of the machine, percent utilization, actual machining time (AMT), and energy consumption of the machine are computed and fed to feedback-based efficiency tracking algorithms \cite{QIU201316}.

\section*{Business Implications}

The above-proposed idea of SIP revolving around SHIP and MCSA has several pros and cons. Business managers need to carefully consider the implications on per case basis. For example, in the context of dynamic storage pricing, which is of central importance to operationalizing SHIP, it is important to realize that it will not always be profitable for SHIP to charge a higher delivery cost \cite{6736130}. Hence, the SHIP must decide on the delivery charge carefully. SHIP is more likely to charge a higher storage cost when it charges a low delivery cost. Small production-scale businesses are likely to rent more storage space from SHIP when the delivery cost or market storage price is low. The SHIP operator should carefully choose the shipping fee to bring in a variety of manufacturers. Businesses with limited production volumes should be aware of the value of market storage prices, SHIP delivery charges, and public warehouse delivery costs because these factors significantly influence the overall expenses. Reducing these fees might result in significant cost savings for small production-scale businesses. \\

Since vehicle size and delivery costs play a vital role in the transportation cost of the SHIP, increasing the capacity of the vehicles can improve the profitability of SHIP but adopting a significantly large-capacity vehicle may not always be beneficial \cite{RePEc:eee:proeco}. Lowering the delivery cost may increase SHIP's profit while not affecting manufacturer performance. The member industries in SHIP should also understand the importance of vehicle capacity and storage costs. These factors significantly impact their delivery/replenishment decisions and overall expenses. The SHIP operator is going to charge a high transportation fee when holding costs are high. In this case, the manufacturers must deal with high holding and transportation costs. Transportation service sharing should be able to bring several benefits to SHIP. However, it depends upon several key parameters such as delivery cost, vehicle capacity, storage price, and the holding cost of the manufacturer. It should be advantageous for the member industries to use shared transportation services. With low storage costs, using large-capacity vehicles may help realize significant cost savings. Enterprises should consider these aspects when deciding whether to use the shared transportation service. \\
    
Now coming to the idea of integrating MCSA in SIP, we may adopt a novel auction-based approach in order to properly utilize the available manufacturing cloud services (MCSs), which maximizes overall profits of customers, providers, and cloud platform operators, thus leading to more participation in MCS trading, decreased wastage of MCSs, and integration of idle MCSs \cite{Kang2022}. This method can be enhanced even more by rolling out logistics resource-sharing services. Cloud platforms should be able to offer a secondary market where customers are authorized to submit bids and requirements to provide logistic resource-sharing services \cite{KANG2021123881}. This service would make better use of idle private logistics assets. To advance sustainability goals, governments need to support such logistics product service systems (LPSS) services as part of SIP ventures.

\section*{Conclusion and Way Forward}

In this article, we aimed to highlight the immense potential that robotic automation and AI algorithms hold for revolutionizing the logistics sector in India. By adopting SIPs and DTs, the Indian logistics sector would be able to make substantial strides toward sustainability by reducing its environmental footprint caused because of transportation emissions, excessive packaging waste, and inefficient resource utilization. DTs would not only help monitor and automate large-scale infrastructures and warehouses but also foster a safer and more comfortable working environment for human-robot collaboration to flourish. DTs would allow businesses to monitor multiple warehouses from remote centralized command centers while making warehouse robots operate 24/7 without the need for breaks or holidays, thus minimizing labor expenses and increasing productivity. It would allow for overall improved inventory management via optimized warehouse layouts, diminished fuel consumption, and reduced inventory holding costs. DTs would help big-sized business units such as large vertical hydroponic farms achieve streamlined and accelerated order processing. By bringing together diverse operations and helping them connect optimally to each other, SIPs promise to deliver maximally in the areas of optimized route planning, dynamically adaptive operations, reduced error rates, handling of fluctuations in demand, reduced transportation costs, and faster delivery times. \\

Since the world is becoming more and more fast-paced and interconnected, we as a nation are rapidly reaching a critical juncture where we need to incorporate all these AI and robotics innovations to transform the logistics sector are critical in today's world. Thus, researchers play a key role in developing and refining these technologies to suit the specific needs and challenges of the Indian logistics industry. It is imperative for researchers, industry stakeholders, and policymakers to collaborate closely, promote investments, and create an enabling environment that fosters innovation and the widespread adoption of such innovations in the logistics sector. By doing so, we may unlock the full potential of these technologies and pave the way for a truly optimized and sustainable logistics ecosystem that will effectively address all the current cost and inefficiency issues plaguing the sector while hopefully bringing down the logistics costs from the current $15\%$ to $8\%$ (refers here to $\%$ of GDP).

\section*{Acknowledgement}

We would like to offer our special thanks to Ph.D. student Muhammad Alfas ST and M.Tech. student Adarsh Mishra for assisting us with a thorough literature review. We would also like to express our great appreciation to Prof. S.K. Gupta from USC and Prof. Amit Kumar from IIT-D whose advice and proofreading have been of much help to us.

\printbibliography

\end{document}